%% file: main.tex
\pdfoutput=1 % arXiv reads this in the first five lines: pdfLaTeX, with the PDF figures
\input{glyphtounicode}\pdfgentounicode=1 % ToUnicode maps for the Type 1 text fonts, as in the template's PDF
\PassOptionsToPackage{table}{xcolor}
\PassOptionsToPackage{breaklinks=true,colorlinks,citecolor=blue,urlcolor=blue,linkcolor=blue,bookmarks=true,bookmarksnumbered=true,bookmarksopen=true,bookmarksopenlevel=2}{hyperref}
\PassOptionsToPackage{noabbrev,nameinlink,capitalize}{cleveref}

\documentclass[onecolumn, numbers]{nanomuse_report}

\usepackage[utf8]{inputenc}
\usepackage[T1]{fontenc}
\usepackage{url}
\usepackage{xurl} % lets the long URLs in the bibliography break anywhere
\usepackage{amsfonts,amssymb,amsmath}
\usepackage{nicefrac}
\usepackage{microtype}
\usepackage{xspace}
\usepackage{fix-cm}
\usepackage{booktabs}
\usepackage{multirow}
\usepackage{makecell}
\usepackage{tabularx}
\usepackage{colortbl}
\usepackage{pifont}
\usepackage{enumitem}
\usepackage{xcolor}
\usepackage{etoolbox}
\usepackage{fancyhdr}
\usepackage{caption}
\usepackage{textcomp}
\usepackage{multicol}
\usepackage{float} % [H] for the figure on page one

\definecolor{aprilblue}{HTML}{0064D4}
\definecolor{linkcolor}{named}{aprilblue}
\definecolor{urlcolor}{HTML}{0064D4}
\definecolor{citecolor}{RGB}{66,168,235}
\definecolor{rowtint}{HTML}{F1F5FA}

\newcommand{\cmark}{\ding{52}\xspace}
\newcommand{\xmark}{\ding{56}\xspace}
\newcommand{\nanomuse}{nanoMuse\xspace}
\newcommand{\muse}{Muse\xspace}
\newcommand{\sentinel}{Sentinel\xspace}
\newcommand{\cny}[1]{\textyen#1}
\newcommand{\file}[1]{\texttt{#1}}
\newcommand{\obs}[1]{\textbf{#1}}
\newcommand{\diagram}[2][\linewidth]{\includegraphics[width=#1]{figures/diagrams/#2.pdf}}
\newcounter{bpar}
\newcommand{\bpar}[1]{\refstepcounter{bpar}\paragraph{#1}\pdfbookmark[2]{#1}{bpar\thebpar}}
\setlist[itemize]{nosep,leftmargin=1.2em,labelsep=0.5em,topsep=2pt}
\setlist[description]{nosep,leftmargin=1.2em,topsep=2pt,font=\normalfont\itshape}

\usepackage[breaklinks=true,colorlinks,citecolor=blue,urlcolor=blue,linkcolor=blue,bookmarks=true,bookmarksnumbered=true,bookmarksopen=true,bookmarksopenlevel=2]{hyperref}
\usepackage{bookmark}
\AtEndPreamble{
    \usepackage[capitalize]{cleveref}
    \crefname{section}{Sec.}{Secs.}
    \Crefname{section}{Section}{Sections}
    \crefname{table}{Tab.}{Tabs.}
    \Crefname{table}{Table}{Tables}
    \crefname{figure}{Fig.}{Figs.}
    \Crefname{figure}{Figure}{Figures}
}
\hypersetup{colorlinks=true,linkcolor=linkcolor,urlcolor=urlcolor,citecolor=citecolor,
  pdftitle={nanoMuse: An Open-Source Personal Agent for Every Device You Own},
  pdfauthor={Guangyi Liu, Yong Liu, Jiangning Zhang}}

\DeclareCaptionFormat{custom}{{\color{aprilblue}\sffamily\textbf{#1 #2}} #3}
\titleformat*{\section}{\color{aprilblue}\Large\sffamily\bfseries}
\titleformat*{\subsection}{\color{aprilblue}\large\sffamily\bfseries}
\titleformat*{\subsubsection}{\color{aprilblue}\normalsize\sffamily\bfseries}
\titlespacing*{\section}{0pt}{1.3ex plus 0.5ex minus 0.3ex}{0.5ex}
\fancypagestyle{firstpage}{\fancyhf{}\fancyfoot[C]{\thepage}}
\patchcmd{\mymaketitle}
  {\includegraphics[width=2.6cm]{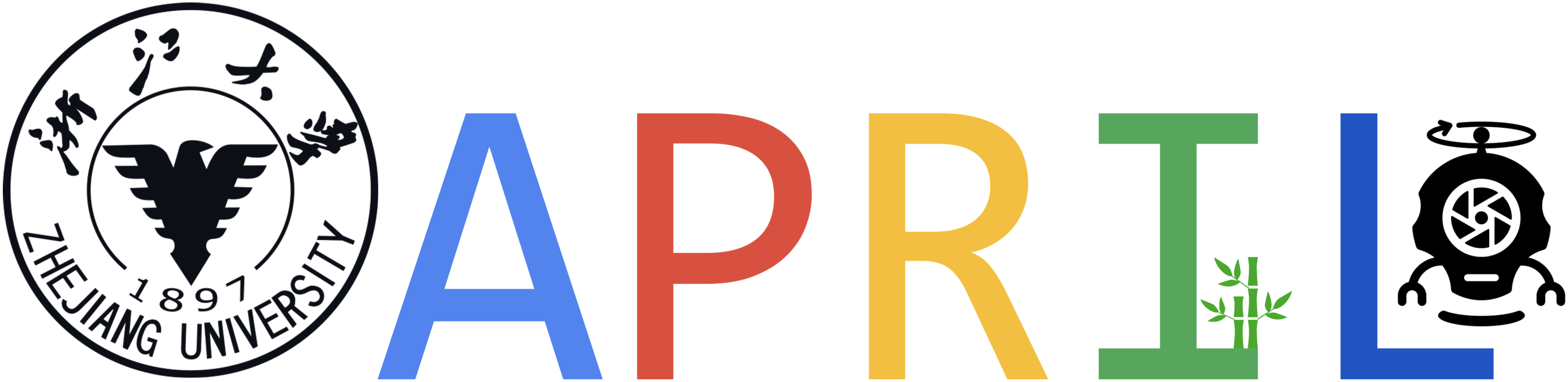}}
  {\includegraphics[height=0.58cm]{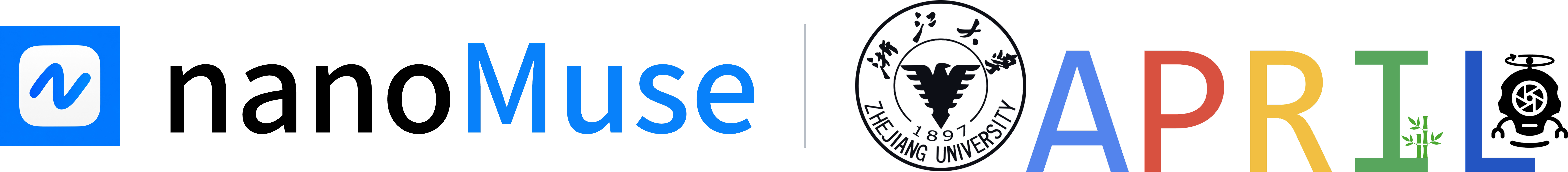}}
  {}{\GenericWarning{}{the title-box logo could not be patched}}
\patchcmd{\mymaketitle}{\setlength{\parskip}{0.5cm}}{\setlength{\parskip}{0.3cm}}
  {}{\GenericWarning{}{the title-box skip above the abstract could not be patched}}
\patchcmd{\mymaketitle}{\abstractlist\par\vskip 0.5cm}{\abstractlist\par\vskip 0.3cm}
  {}{\GenericWarning{}{the title-box skip below the abstract could not be patched}}

\renewcommand{\title}[1]{\def\titlelist{\parbox{\linewidth}{\raggedright
  {\linespread{1}\fontsize{20pt}{24pt}\selectfont\sffamily\bfseries #1\par}\vspace{3pt}}}}
\title{nanoMuse: An Open-Source Personal Agent for Every Device You Own}

\author[1]{Guangyi Liu}
\author[1]{Yong Liu}
\author[1,\dagger]{Jiangning Zhang}

\affiliation[1]{Zhejiang University}

\contribution[\dagger]{Corresponding author}

\abstract{\input{sec/0_abstract}}

\coverdate{October 2026}
\covercorrespondence{\email{186368@zju.edu.cn}}
\coversourcecode{https://github.com/nano-muse/nanoMuse}
\coverproject{https://nanomuse.cn}
\coverdemo{https://nanomuse.cn/web/}

\begin{document}

\maketitle
\thispagestyle{firstpage}

% Figure 1 sits under the abstract, on page one, in place of a table of contents (the PDF bookmarks are the
% contents). Placed here, not floated, so that it cannot drift.
\begin{figure}[H]
  \vspace{-12pt}
  \centering
  \diagram[0.96\linewidth]{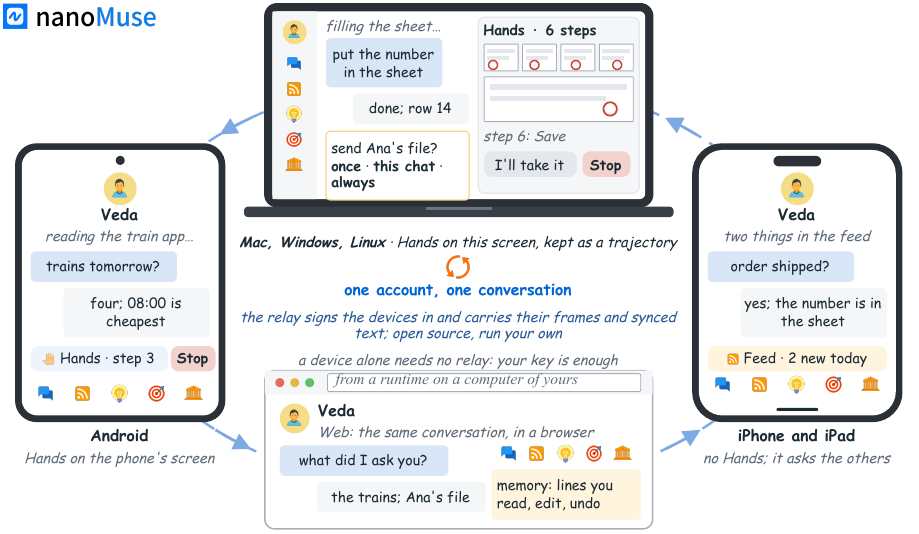}
  \caption{\textbf{One agent on every device, one conversation between them.} Android and the desktop carry the whole agent and have hands on their own screens; the iPhone has everything but the hands, since iOS lets no app operate another; the web app shows the same conversation from a runtime on a computer of the person's own. The devices meet over the relay, open source and optional; the screens are drawings, not screenshots.}
  \label{fig:devices}
  \vspace{-\intextsep}% nothing follows on this page; the skip would push the figure off it
\end{figure}
\clearpage% so that Section 1's anchor lands on its heading's page, not at the foot of page one

\input{sec/1_why}
\input{sec/2_sofar}
\input{sec/3_muse}
\input{sec/4_should}
\input{sec/5_nanomuse}
\input{sec/6_roadmap}
\input{sec/7_limitations}

% fifty short entries in two columns keep the card inside twelve pages
\setlength{\premulticols}{3\baselineskip}
\setlength{\multicolsep}{0pt}
\begin{multicols}{2}[\section*{\refname}\phantomsection\addcontentsline{toc}{section}{\refname}]\raggedright\linespread{0.92}\selectfont
\bibliography{refs}
\end{multicols}

\end{document}

%% file: sec/0_abstract.tex
Assistants from 2011 answered and waited, and agents from 2023 did a task and stopped. In September 2026 Meta's \muse showed an agent for one person, with accounts, devices, memory and a conversation that lasts, closed, in a vendor's cloud, in one country. Such an agent is expected to act on a person's accounts and devices, remember them across weeks, speak first when it is worth it, and answer for what it did. It is a kind of software, not a model, and until now had no open counterpart. This report defines the personal agent in five questions and three horizons. It reads how \muse is built from Meta's public record and a copy of its production prompt, each statement marked by its source. It then presents \nanomuse, the open-source counterpart under the GPL-3.0, one agent on every device a person owns, with hands on the phone's screen and the computer's. They share one conversation over a relay anyone can run; every action goes through a \sentinel, memory is files the person can read, and the model is their choice. Its size and cost are given as estimates. What is open, memory with provenance, an evaluation suite for the hands and an open model for them, is set out as a roadmap.

%% file: sec/1_why.tex
\section{Introduction}
\label{sec:why}

A personal agent is a program that acts for one person on their accounts, their devices and their files. It works for weeks, partly while they are away, and it answers to them afterwards for what it did, with a log, a ledger and a list of the permissions it holds. Who runs that program, where, and under whose eyes is not a detail of deployment. It is the product.

In September 2026 Meta released \muse, and within three weeks OpenAI, Manus and a new company called Today released agents of the same kind~\cite{meta2026intro,openai2026dots,manus2026cue,today2026what}. \muse showed the field what the product is. It is one agent with a name and one long conversation, with memory the person can read, messages it sends first, a wallet behind an approval, and a safety architecture written up in public~\cite{sheasha2026safety}. It also runs in one Linux VM per person in Meta's cloud and is sold in the United States only. Nothing of it is open except a small SDK, published on October 2, for the gadgets it can talk to~\cite{musegadget2026sdk}.

We think the second half of that description is the wrong shape for the first half. An agent that holds a person's life should be inspectable by that person or by anyone they trust. It should run on the hardware they already own where that is possible, and on a server they chose where it is not. The one piece that has to be shared, the meeting point of their devices, should be code anyone can read and run. This is not an argument against \muse. It is an argument that the open counterpart has to exist, as a complete thing a person can run. \nanomuse is our attempt at that counterpart, and \cref{fig:devices} is the shape of it.

\bpar{Contributions.} (1) A definition of the personal agent opens the report, with the history from the assistants of 2011 to the agents of September 2026 (\cref{sec:sofar}). (2) A reading of how \muse is built follows, from Meta's public record and a copy of its production prompt, each statement marked by its source (\cref{sec:muse}). (3) A statement of what a personal agent should be comes as five questions and three horizons (\cref{sec:should}). (4) \nanomuse, an open personal agent on every device a person owns, is presented with hands on the phone's screen and the computer's, a \sentinel over every action, memory as files, and its size and cost in numbers (\cref{sec:nanomuse}). (5) A roadmap of the open work closes the report, with what we would not yet trust it with (\cref{sec:roadmap,sec:limits}).

%% file: sec/2_sofar.tex
\section{Personal Agents So Far}
\label{sec:sofar}

\begin{figure}[t]
  \centering
  \diagram[0.96\linewidth]{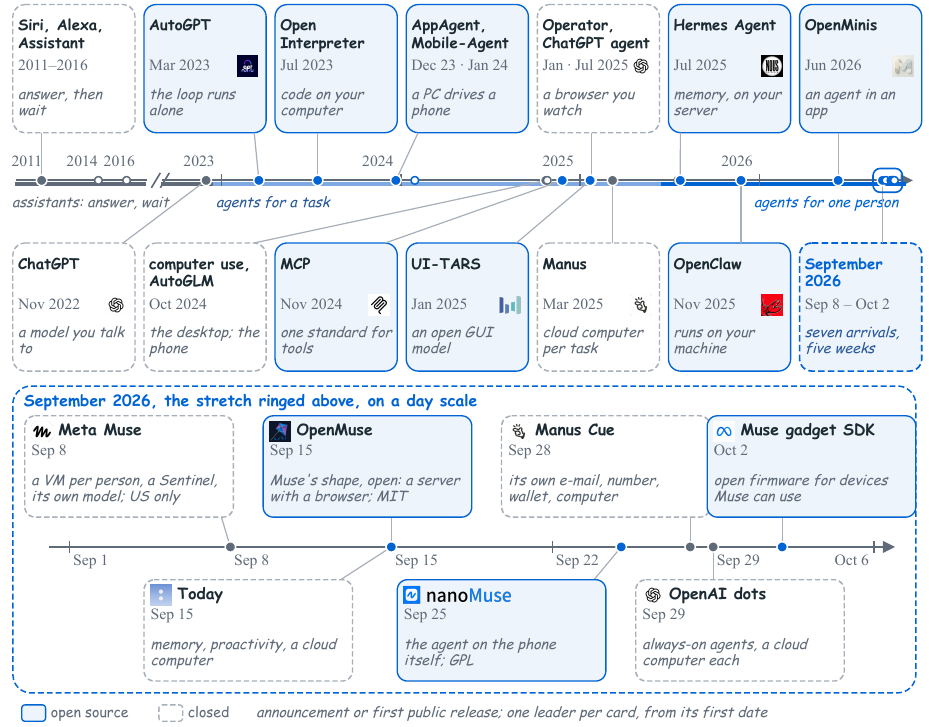}
  \caption{\textbf{From assistants to agents, 2011--2026, and the five weeks in which personal agents arrived.} Above, a broken axis with a dot per date and three eras under it; below, September 2026 on a day scale. Blue cards are open source, white dashed cards closed; logos are the makers' marks.}
  \label{fig:timeline}
\end{figure}

\bpar{Assistants, then agents.} \Cref{fig:timeline} is the short history. Siri, Alexa and Google Assistant answered a question and waited for the next. ChatGPT could be told to do things, and within months people had it doing them: AutoGPT let the loop run unattended~\cite{autogpt2023}, Open Interpreter let a model write and run code on the user's own computer~\cite{openinterpreter2023}. The screen became a tool next: CogAgent trained a vision-language model to read one~\cite{hong2024cogagent}, AppAgent and Mobile-Agent drove a phone from a PC through its screenshots~\cite{zhang2023appagent,wang2024mobileagent}, Anthropic's computer use drove a desktop the same way~\cite{anthropic2024computeruse}, Zhipu's AutoGLM drove a phone through its accessibility tree~\cite{liu2024autoglm}, and MCP gave agents one standard for tools~\cite{anthropic2024mcp}. Open models that operate a screen followed, OS-Atlas, Aguvis and UI-TARS~\cite{wu2024osatlas,xu2024aguvis,qin2025uitars}, then OpenCUA and Mobile-Agent-v3's GUI-Owl~\cite{wang2025opencua,ye2025mobileagentv3}, measured on live environments: Mind2Web and WebArena for the web~\cite{deng2023mind2web,zhou2023webarena}, OSWorld for the desktop~\cite{xie2024osworld}, AndroidWorld for the phone~\cite{rawles2024androidworld}; the phone side of this history is surveyed in~\cite{liu2025survey}. Then the vendors moved the agent into their own clouds: Operator~\cite{openai2025operator} and ChatGPT agent~\cite{openai2025agent} were hosted browsers the user watched, Manus a cloud computer per task~\cite{manus2025}. Every one of these acts for a task; none belongs to a person. Two open projects crossed that line from the other side: Hermes Agent and OpenClaw run on a computer or server of the person's own, remember across sessions, and are reached through the chat apps the person already uses~\cite{hermes2025,openclaw2025}; neither has a screen. OpenMinis (June 2026) is the opposite construction: a complete agent inside an Android app, GPL, no server~\cite{openminis2026}.

\bpar{September 2026.} The lower half of \cref{fig:timeline} expands five weeks, and \cref{tab:wave} lists what arrived in them. \muse (September 8) is the first product to assemble the pieces for one person and to say so in its first sentence~\cite{meta2026intro}. Today (September 15) is a personal agent from a new company, with memory, proactive follow-up and a cloud computer~\cite{today2026what,today2026meet}. The same day CopilotKit published OpenMuse, open source and in \muse's shape: a server with a persistent browser, the apps as windows onto it~\cite{copilotkit2026openmuse}. Manus Cue (September 28) gives each agent its own e-mail address, phone number, wallet and computer~\cite{manus2026cue}; OpenAI's dots (September 29) are ``always-on agents'' with a cloud computer each~\cite{openai2026dots}. On October 2 Meta opened one edge of \muse: a gadget SDK under the Apache licence, so that a device a person builds can pair with their \muse~\cite{musegadget2026sdk}; the agent, the service and the model stay closed. Four companies, five weeks, one architecture: a computer per person in the vendor's cloud, thin clients on the person's devices, a model the vendor trains.

\begin{table}[t]
  \centering
  \small
  \caption{\textbf{The agents of September 2026, and the open projects they are measured against.} ``Where it runs'' is where the agent's own loop executes; a screen column means it can operate other applications through that device's display. Dates are announcements or first public releases.}
  \label{tab:wave}
  \setlength{\tabcolsep}{4pt}
  \renewcommand{\arraystretch}{1.0}
  \footnotesize\begin{tabularx}{\linewidth}{@{}>{\raggedright\arraybackslash}p{4.2cm}>{\raggedright\arraybackslash}p{1.8cm}>{\raggedright\arraybackslash}X>{\centering\arraybackslash}p{1.1cm}>{\centering\arraybackslash}p{1.4cm}>{\centering\arraybackslash}p{1.3cm}@{}}
    \toprule
    System (maker) & Since & Where the agent runs & Phone screen & Computer screen & Open \\
    \midrule
    Hermes Agent (Nous Research) & Jul 2025 & a computer or server of yours & \xmark & \xmark & MIT \\
    OpenClaw (OpenClaw Foundation) & Nov 2025 & a computer or server of yours & \xmark & \xmark & MIT \\
    OpenMinis (its contributors) & Jun 2026 & inside the Android or iOS app & \cmark\textsuperscript{a,d} & \xmark & GPL-3.0 \\
    \muse (Meta) & Sep 8, 2026 & one Linux VM per person, Meta's cloud & \xmark & \cmark\textsuperscript{b} & no\textsuperscript{f} \\
    Today (Today) & Sep 15, 2026 & a cloud computer, with local apps & \xmark & \cmark\textsuperscript{c} & no \\
    OpenMuse (CopilotKit) & Sep 15, 2026 & a server of yours: a browser and a Linux workspace & \xmark & \xmark\textsuperscript{e} & MIT \\
    Manus Cue (Manus) & Sep 28, 2026 & a cloud computer per agent & \xmark & \xmark & no \\
    OpenAI dots (OpenAI) & Sep 29, 2026 & a cloud computer per agent & \xmark & \xmark & no \\
    \nanomuse (this project) & Sep 25, 2026 & the phone and the computer; relay optional & \cmark\textsuperscript{a} & \cmark & GPL-3.0 \\
    \bottomrule
  \end{tabularx}
  \par\smallskip
  \raggedright\footnotesize
  \textsuperscript{a}~Android only; iOS lets no app operate another, so no agent has a phone screen there.\quad
  \textsuperscript{b}~Through the \muse macOS app, as a dimmed live stage with a stop control.\quad
  \textsuperscript{c}~Today's own description; not verified by us.\quad
  \textsuperscript{d}~An accessibility executor; the screenshot loop of \cref{sec:nanomuse} is \nanomuse's.\quad
  \textsuperscript{e}~OpenMuse's own description; graphical desktops are listed as future work.\quad
  \textsuperscript{f}~The gadget SDK of October 2, 2026 is Apache-2.0; the agent, the service and the model are not open~\cite{musegadget2026sdk}.
\end{table}

\bpar{What changed.} Models learned to call tools and to read screens, which made ``act'' possible; memory and background execution made ``for one person'' possible. And the phone is where the person's life is, which makes ``where it runs'' the question that decides the rest: the cloud computers above reach anything with an API or a web page and nothing else, and many of the apps a person uses every day, from banking to government services, have neither. That gap is what \nanomuse's hands are for.

%% file: sec/3_muse.tex
\section{How \muse Is Built}
\label{sec:muse}

\begin{figure}[t]
  \centering
  \diagram[0.96\linewidth]{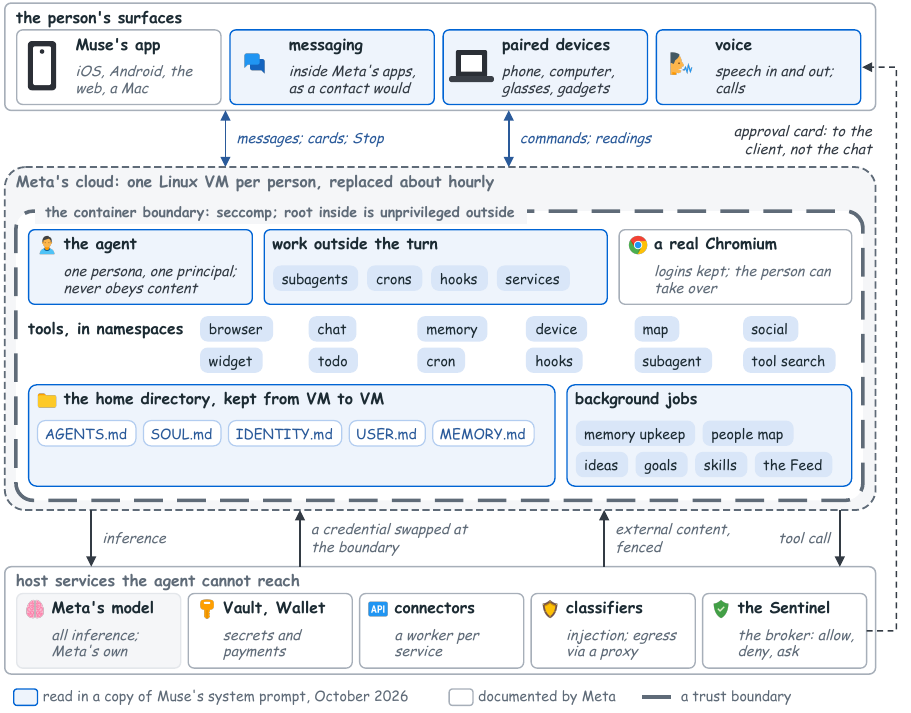}
  \caption{\textbf{\muse as a system.} Three zones from Meta's safety essay: the person's surfaces, Meta's cloud with the container boundary drawn, and the host services the agent cannot reach; the arrows are the flows across each boundary. Blue tint: read in a copy of \muse's production system prompt, October 2026; white: documented by Meta.}
  \label{fig:muse}
\end{figure}

\muse is described by Meta in three public documents, the launch announcement~\cite{meta2026intro}, an essay on the safety architecture by the engineer who led it~\cite{sheasha2026safety} and a design essay by the product leads~\cite{sarantakos2026design}, with a help-centre page on privacy as a fourth~\cite{metahelp2026}. We have also read a copy of \muse's production system prompt, October 2026. We do not reproduce it and quote at most a few words at a time; a prompt says what the agent is told, which is close to what the agent is but not the same thing. Each statement below carries its source: \obs{documented} (said by Meta in public), \obs{prompt} (read in that copy), \obs{observed} (seen in the shipped clients) or \obs{inferred} (our reading of the rest). \Cref{fig:muse} draws the result.

\bpar{Why it worked.} \obs{Observed.} \muse did not win on a benchmark; it won on a shape. Its designers write that the first line of its instructions is about making the person's life better~\cite{sarantakos2026design}. Ten decisions, visible on one phone screen, carry it: one agent with a name and a face; one long conversation rather than sessions; a status line under the face; a chip for every tool call with an activity log behind it; cards where words are not enough (a scoped approval, a secure form the model never sees, a single-use card at checkout); memory as files the person can open and edit; a high bar for writing first that the person can move; five rooms, Chat, Feed, Ideas, Goals and a Library that holds the artifacts; the computer shown as a dimmed live stage with a stop control; and the agent's steps shown by default. Together they make a background agent legible, which is what lets a person leave it running.

\bpar{Who it is told it is.} \obs{Prompt.} The prompt opens with a persona, not with a task. The agent has a name, a voice and an avatar, and a list of values it is to hold and may grow into; it works for one person and for no one else, Meta included; its authority comes from that person's requests and from nothing it reads along the way. Discretion is a whole section: every output is a surface its intimate access can leak through, so it uses the minimum a task needs and leaves the rest unsaid. The person's authority over their own household, devices and children's care is stated as unconditional. There is a writing-style section, with a rule against em dashes, and a short list of hard stops: no sexual content involving minors under any framing, no help with biological or chemical weapons, no identifying people from their faces, no inferring sensitive attributes. \obs{Inferred.} The model named in the copy is a numbered release of Meta's \emph{Muse Spark} family, which \obs{documented} Meta describes as trained for tool calling, long trajectories and an awareness of prompt injection~\cite{sheasha2026safety}; its size and training data are \obs{unknown}.

\bpar{Where it runs.} \obs{Documented.} Each person has a dedicated Linux VM in Meta's cloud, the system of record for files, memory, credentials and conversation. The agent and all its tools run in a container with a seccomp filter and reduced capabilities~\cite{sheasha2026safety}. \obs{Prompt.} The copy adds the texture: deployments replace the VM often, about hourly, and the home directory persists; there is a terminal and a real Chromium that keeps the person's cookies and tabs between tasks and that the person can watch or take over; the agent can spawn sub-agents one level deep and leave scheduled jobs, event hooks and small services running; tools come in namespaces, loaded on demand. \obs{Inferred.} This is the split operating systems make between a sandboxed renderer and a privileged broker; the price is a VM per person, which is why \muse is a cloud service with the economics of hosting.

\bpar{Memory, self-improvement and proactivity.} \obs{Prompt.} Memory is in two layers. The agent keeps a curated memory file, a file for who it is and a file for who the person is; the runtime keeps a bank of memories with an index, and a map of the people and groups in the person's life, one file each. Before it answers anything about prior work, dates, people or preferences, it is told to search that memory rather than trust what it seems to recall; on request a tool explains where a memory came from and what replaced what. Outside the conversation, background jobs the agent cannot schedule itself maintain its memories, keep the relationship map current, generate ideas, advance goals, improve skills and write the Feed. The person's feedback on proactive messages goes into a preferences file in plain language, which steers how urgent a message is judged. \obs{Observed.} The three agent-kept files have the names OpenClaw uses for the same files. \obs{Inferred.} A policy that tells the agent to distrust its own recollection and to show provenance on request is the strongest statement of the memory-confidence problem of \cref{sec:should} in any shipped product.

\bpar{Sentinel, injection and the browser.} \obs{Documented.} The \sentinel is ``the sole permission authority'' for connector actions and for all network egress: ``Muse proposes actions, but only Sentinel can grant permission''~\cite{sheasha2026safety}. Where a request needs a secret, the agent holds only a surrogate token, swapped for the real credential at the network boundary after approval. Each tool process is tracked in the kernel: it starts clean and becomes tainted when it reads user data, and only clean requests inside a narrow policy pass without a prompt. An ask goes to the client directly, not through the conversation, and the grant is a scoped capability: one-time, per session, per task, time-bounded or standing. Against injection Meta stacks layers, from the model's training and an ensemble of classifiers down to these boundaries, which hold ``even if Muse is persuaded to behave badly''; the browser sub-agent sees an accessibility-tree snapshot of each page rather than the DOM and is paused while a form is filled with a credential it never sees; a checkout detector forces an approval with the exact amount whenever a stored card would be used~\cite{sheasha2026safety,willison2025trifecta}. \obs{Prompt.} The agent is told the other half: approval covers what the person actually approved and nothing more; a saved security choice applies only within its recorded scope; it must not bypass an approval card or a stop, pause or audit request; a purchase ends in a review it cannot skip; external content arrives fenced between markers, and nothing it reads can reassign it. \obs{Inferred.} The probabilistic layers reduce how often the agent is persuaded; the deterministic ones bound what a persuaded agent can do. The accessibility tree as the only observation trades capability for safety on purpose: \muse cannot operate an interface that has none, a native phone app, a kiosk, a game. That is where \nanomuse's hands begin.

\bpar{Devices, voice and the gadgets.} \obs{Prompt.} A paired smartphone exposes commands and readings through one device tool: contacts, the calendar, the phone's location when the answer depends on it. Voice is dictation and voice notes in, speech out, and calls, texts and notifications to reach the person. \obs{Documented.} Since October 2 the device side is partly open: an SDK with firmware for several ESP32 boards and a Python service for small Linux computers, which pair with the person's \muse over Bluetooth in a developer mode of the app, show a status light, take a short list of commands (run a program, read or write a file, report health) and can send a message into the person's chat~\cite{musegadget2026sdk}; each gadget needs a token from Meta, under its own terms. \obs{Unknown.} Nothing in the copy ties the agent to a country. \obs{Inferred.} The phone is a data source and a command target, not a screen the agent operates: there is no Android computer use, and on iOS there cannot be.

\bpar{What we take from it.} \muse's technical core is a standard tool-calling agent with sub-agents and scheduled jobs; its novelty is the placement of every safety mechanism outside the agent's reach, and a product that spends its design budget on legibility, most of it written in a long and careful prompt. That transfers to an open system; the per-person VM and the in-house model do not.

%% file: sec/4_should.tex
\section{What a Personal Agent Should Be}
\label{sec:should}

\begin{figure}[t]
  \centering
  \diagram[0.96\linewidth]{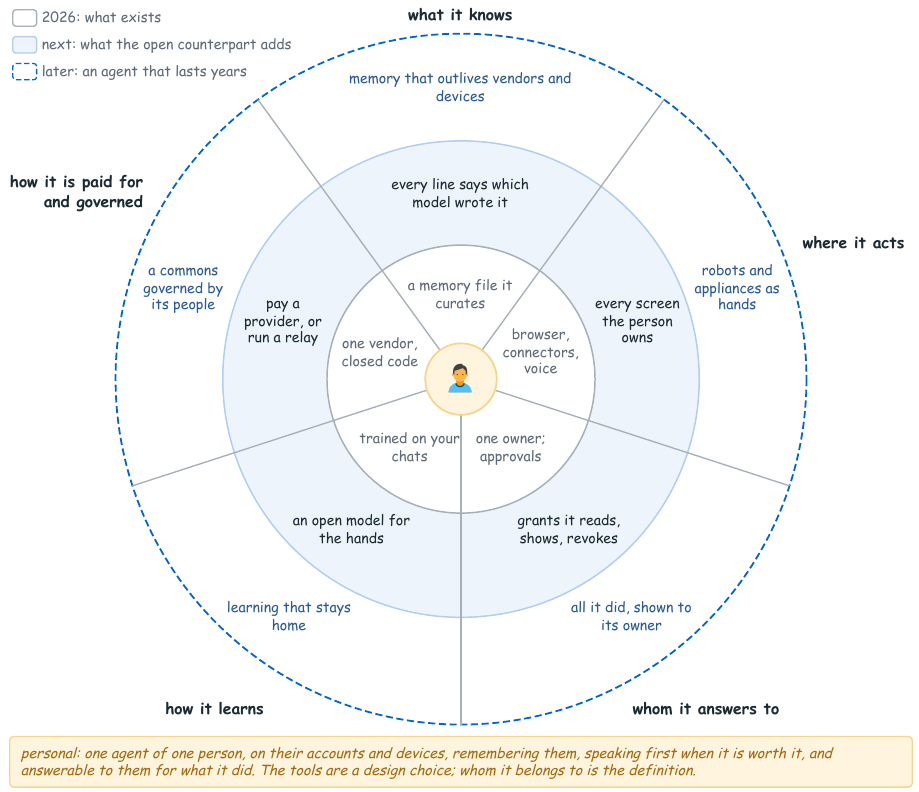}
  \caption{\textbf{What a personal agent should be: the person at the centre, five questions as sectors, three horizons as rings.} Voice sits in the inner ring on purpose: assistants have spoken since 2011 and \muse speaks and listens today.}
  \label{fig:insight}
\end{figure}

This section is opinion, meant to reach past anything launched this autumn, ours included: five questions about the agent, each answered for what exists in 2026, for what the open counterpart should add next, and for an agent meant to last years. \Cref{fig:insight} draws them; the paragraphs below are its sectors.

\bpar{What it knows: memory the person can read, edit and take away.} An agent that remembers a person for years will hold the most complete record of them that exists anywhere. That record should be plain files, in the person's language, theirs to open, correct, delete and export in a form another agent can read; it is what lets the agent outlive its models and devices. Memory needs provenance: each line should carry the model that wrote it, the date, and how sure it was, because a line written by a small cheap model one week will be read by a stronger one the next. It needs a reconciliation rule: when a new question overlaps an old line, the agent of the day re-checks it against what it now sees and confirms, updates or doubts it, recording which. A line the person wrote outranks any line a model wrote.

\bpar{Where it acts: one agent, many hands.} A person should have one agent, not one per app or one per device, and it should be their standing representative: the thing that acts for them when they are not there, under their name, on their accounts. Everything a person owns that has a screen, a microphone or a motor is then a pair of hands for that one agent and a door through which the person reaches it. In 2026 the hands are a cloud computer with a browser, the connectors, the voice and a few gadgets; the phone's own screen only in \nanomuse. Next come the screens the person already owns; after that robots and appliances, since a robot that does not share its owner's agent is an appliance.

\bpar{Whom it answers to: approvals as the grammar of trust.} A personal agent will do things that cannot be undone: send, pay, delete, sign. Granting and withdrawing the right to do them is how a person talks to the agent about trust, and it should be a small, regular language: a grant with a scope the person can say in a sentence (once; for this conversation; always, for this recipient, this site, this folder), listed where the person can read and revoke it, asked for on whichever device the person is holding, with the thing being approved shown as it will be done. Some grants should not exist: a payment approval or a password is never remembered, however often the person says ``always''.

\bpar{How it learns: in the open, with consent.} The difference between an agent and an assistant is that the agent speaks first, and that is also where it will lose a person first. The bar for interrupting should be something the person can see and move, and a feed of what happened while the person was away should read like a note from a competent colleague. The hard version, which no one has built, learns the person's attention well enough to time its interruptions, and asks first. The other half of learning is the model: in 2026 the vendor trains it on the chats unless the person opts out. The open counterpart should train the one model that touches the most sensitive loop, the hands, on traces people chose to share, the way a few demonstrations teach a phone agent an app it has not seen~\cite{liu2025learnact}, and measure it with a suite anyone can run; later the learning should stay home, the person's own devices improving their agent without sending their life anywhere.

\bpar{How it is paid for and governed: small enough to run at home.} An agent that costs a VM per person is paid for by a subscription or a vendor's other business, and whoever pays decides. The open counterpart should let the person pay a provider directly, or run the meeting point of their devices for the price of the smallest server, with code and data formats they own. In the long run a personal agent is a commons: skills, evaluations and an open model governed by the people whose data it is; a non-profit project with a public repository is the smallest start. \Cref{sec:nanomuse} is measured against these five questions and falls short of most.

%% file: sec/5_nanomuse.tex
\section{\nanomuse Today}
\label{sec:nanomuse}

\begin{figure}[t]
  \centering
  \diagram[0.96\linewidth]{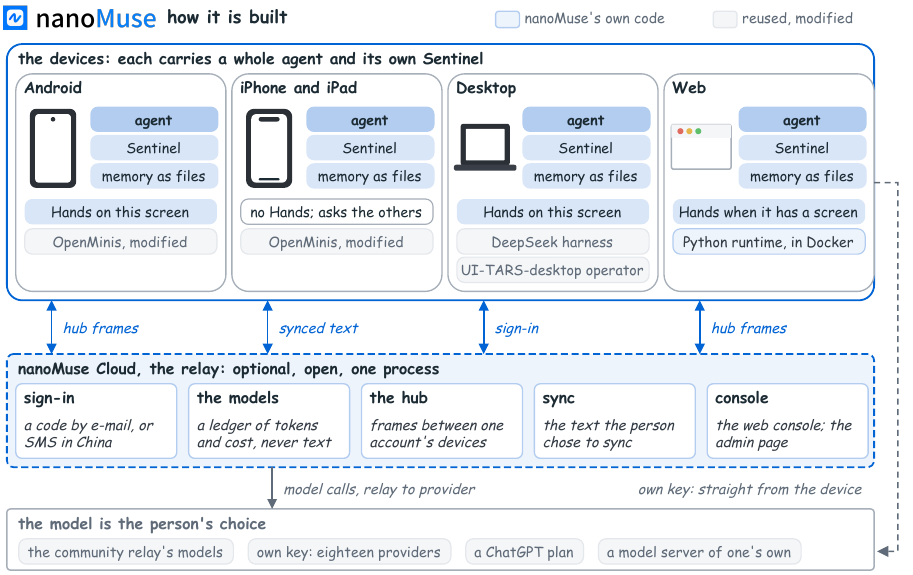}
  \caption{\textbf{How \nanomuse is built, as of version 0.1.40.} Each device is a whole agent with its own \sentinel; the relay is the one shared piece, optional and open; with a key of its own a device talks to the provider directly, past the relay.}
  \label{fig:arch}
\end{figure}

\nanomuse is a minimal, complete personal agent under the GPL-3.0: an Android app, an iOS app (TestFlight), a desktop app for macOS, Windows and Linux, a web app, and the relay, \emph{nanoMuse Cloud}, with its console. The first release was on September 25, 2026; this report describes the code as of version 0.1.40 (October 2026).\footnote{Later versions will differ; the changelog is the record.} ``Minimal'' is meant literally: every piece exists and works end to end, and most pieces are the simplest thing that does. \Cref{fig:arch} is the construction.

\bpar{Where it differs.} Against \muse, \nanomuse adds hands on the Android phone's screen; the phone and the computer are operated on the device itself, not in a VM; every part is open, the relay included; and the model is the person's choice. It lacks a model of its own, the per-person VM with its kernel-level taint tracking and surrogate credentials, the wallet, the artifacts, and the polish of a large team. Against OpenMinis, the phone agent it is built from~\cite{openminis2026}, it adds everything that makes an agent personal, from the face to the relay. Against CopilotKit's OpenMuse, open but built as a server with windows onto it~\cite{copilotkit2026openmuse}, \nanomuse puts the agent on the device, which gives it the phone's screen and lets one device work with no server.

\bpar{Devices.} The Android app is OpenMinis, modified: a complete agent inside the app, with a sandboxed Alpine Linux, a shell, a browser, MCP servers, skills and scheduled tasks, talking to any OpenAI-compatible model; \nanomuse's additions sit beside the upstream code so that its releases can still be merged. The desktop app is an Electron shell around the DeepSeek harness~\cite{deepseek-harness2026} with a \nanomuse plugin, carrying the Python runtime inside it; the same runtime serves the web app from a computer of the person's own.

\bpar{Sentinel.} The agent never executes a tool. Every call goes through a gate with a fixed decision order: denied tools; the person's own rules; the always-allow and always-ask lists; \emph{taint}, which once a conversation has read private data turns any call that sends data outside the allow-list into an ask, and never the reverse; the tool's risk against the mode the person set; and last a few warnings (a destructive shell command, a piped download executed on the spot, a committing word in a label) that no mode and no list can wave through. An approval is a scoped grant rather than a yes: \emph{once}, \emph{this conversation} and, where a target can be named, \emph{always} for that recipient, site, folder, device or application, listed under \emph{Permissions} with a revoke button. A call with a warning gets \emph{once} only, so a payment or a password is never remembered; every decision goes into a log the person can open. The \sentinel here is a policy boundary in the same process family, not \muse's privilege boundary (\cref{sec:limits}).

\bpar{Hands.} To reach an app the agent climbs a ladder and takes the lowest rung that works: a skill, a command-line tool or an MCP server; a page fetched with the person's login; the browser; and, last, the device's own screen. This is the hybrid of shortcuts and screen that MAS-Bench measures for phone agents~\cite{zhao2025masbench}. The last rung is \emph{Hands}, off by default: a one-action-per-screenshot loop in the dialect of MemGUI-Bench's phone operator~\cite{liu2026memgui}, with the computer side ported from UI-TARS-desktop~\cite{uitarsdesktop2025}. The model sees one screenshot, the goal, a one-sentence history per step (the record of its own run that a long task on a phone turns on~\cite{liu2026memguiagent}) and the accessibility elements when the device reports them, and answers with a thought, a sentence in the person's language saying what it is about to do, and one gesture. The sentence is what makes this rung governable: a deterministic policy cannot judge a picture, but it can judge the model's words for what it is about to press. A label with a committing word (confirm payment, transfer, place order, send, delete) makes the step sensitive and approved every time; a typing step that submits in one go is sensitive, because Enter in a messenger is a send; a step without a label carries a warning. On Android the accessibility text where the finger will land is read too, and the stricter of the two decides; the operator's prompt forbids typing passwords, PINs, card numbers and one-time codes. On the phone the person sees a capsule over the app with the face, the step and a red \textbf{Stop}, and a ring where the next tap will land. On the computer a glow breathes along the edges of the screen while the hands work, a marker shows where the next click lands, and the run is kept in the chat as a trajectory, each step's screenshot with its action drawn on it, with \emph{I'll take it} to pause the agent and use the mouse. We built \muse's dimmed live stage and dropped it; a trajectory can be looked back at, a moving picture cannot.

\bpar{Memory as files.} On the phone, memory is Markdown files the person can open in the app and the agent can open in its shell: who the agent is (\file{SOUL.md}), who the person is (\file{USER.md}), what holds across conversations (\file{GLOBAL.md}), a dated diary, and the agent's routines (\file{HEARTBEAT.md}). On the computer the store keeps each memory as a line with a change log and an undo, recalled by rare words or, with an embedding endpoint, by meaning; a tidy-up pass proposes merges and drops and never drops what the person wrote. The rooms, Feed, Ideas and Goals, are written by the agent in hidden conversations and read by the person; the feed's bar for interrupting is a setting the person moves.

\bpar{Every device.} \muse's clients are windows onto one agent; \nanomuse's devices are peers, each with its own agent, meeting over the hub on the relay. Any device's agent has the others as tools, so ``find the order number on my phone, then put it in the spreadsheet on my Mac'' is one conversation. The device that would act agrees first, with a grant the person there gives once or for that device, and a job handed to another device runs under that device's own \sentinel, so the asking device cannot lend a permission it does not have.

\bpar{The model is the person's choice.} One device alone needs no server: the person puts an API key of their own into the app and nothing of theirs leaves the device except the model calls~\cite{nanomuse2026ownkey}. The apps and the relay share one catalogue of eighteen providers, Chinese and international, each entry saying what a key from that provider covers (chat, vision, image, video); the one suggested first depends on where the person is. A ChatGPT plan can stand in for a key, signed in through OpenAI's own Codex authorisation flow; it covers chat and the hands, not images or video, and it rests on OpenAI's terms, so a key remains the dependable route. A model server on the person's own machine works too.

\bpar{What the relay keeps.} Working across devices needs the relay, the one shared piece, so we state what it holds~\cite{nanomuse2026privacy}: the account (a hashed identifier, the keys it issued and the device each went to), the ledger of model calls (model, token counts, cost; never the content), the agent's profile, the hub's presence, and the text of the conversations the person chose to sync, the main conversation by default and side chats on request; never files, images or what a tool returned. A synced conversation remembers the account it belongs to, so that two people who share a device each see their own. One switch in \emph{Data controls} turns sync off and deletes what was stored; deleting the account deletes all of it, and the relay keeps only a hash of the revoked keys for ninety days, so that a device that was offline learns the account is gone and nothing else. A second switch, \emph{Help improve nanoMuse's AI models}, keeps the text of chats with the community's models for training an open model; it is on for new accounts on the community relay, says so beside itself, and is off in one tap; a self-hosted relay chooses its own default.

\begin{table}[t]
  \centering
  \small
  \caption{\textbf{What ``nano'' weighs and costs.} Sizes are the release files and one installation, memory one Linux workstation; costs are estimates from the project's documentation and the providers' price lists; nothing here measures anyone's use.}
  \label{tab:nano}
  \setlength{\tabcolsep}{5pt}
  \renewcommand{\arraystretch}{1.0}
  \footnotesize\begin{tabularx}{\linewidth}{@{}>{\raggedright\arraybackslash}p{3.1cm}>{\raggedright\arraybackslash}X>{\raggedright\arraybackslash}p{3.3cm}@{}}
    \toprule
    What & Number & How we got it \\
    \midrule
    Android app & 38 MB to download (arm64) & the release file \\
    Desktop app & 256 MB (Linux) to 498 MB (macOS) to download, 266 MB on Windows; 928 MB installed on Linux; about 0.5 GB of memory idle, all processes together & the release files; one installation; proportional set size here \\
    Relay & one Python process and one SQLite file; about 85 MB of memory idle with a fresh database; sized for a server with 1 vCPU and 1 GB~\cite{nanomuse2026selfhost} & measured here; the self-hosting page \\
    A relay of your own & on the order of \cny{30}--\cny{60} a month inside mainland China or US\$4--6 outside it for the smallest server tier, plus model usage~\cite{nanomuse2026selfhost} & the self-hosting page (estimate) \\
    Model calls & Bailian's October 2026 list: \cny{2} in and \cny{8} out per million tokens for the chat model in busy hours, \cny{3} and \cny{12} for the hands model; a day of chat is a few fen~\cite{nanomuse2026ownkey} & the provider's price list (estimate) \\
    \bottomrule
  \end{tabularx}
\end{table}

\bpar{Running it yourself.} \Cref{tab:nano} is what the small thing weighs and costs. There are three ways to run it~\cite{nanomuse2026selfhost}: one device with your own key or a ChatGPT plan, no account and no server; several devices on the community relay, signed in with an e-mail address or, in mainland China, a phone number; or a relay of your own, in Docker on the smallest server tier, brought up with TLS by one script, everyone who signs in on it billed to your provider key.

%% file: sec/6_roadmap.tex
\section{A Roadmap With the Community}
\label{sec:roadmap}

\begin{figure}[t]
  \centering
  \diagram[0.96\linewidth]{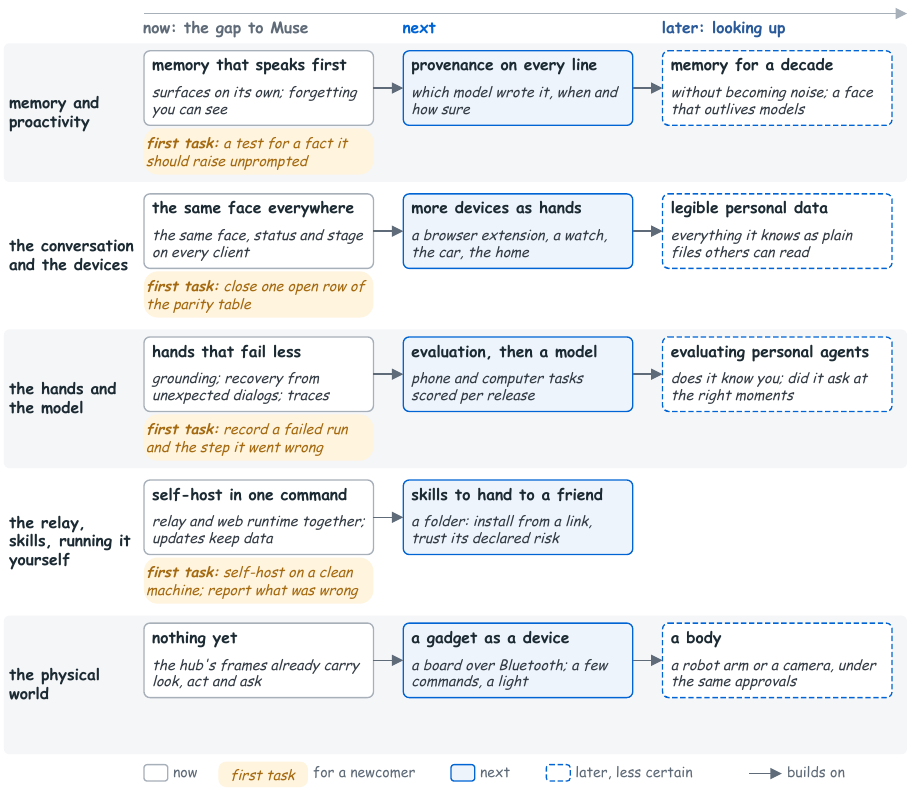}
  \caption{\textbf{The roadmap: five lanes, three distances, one milestone per cell.} Now is the gap to \muse, each card with a first task a newcomer can take; later (dashed) is what no shipped product has; no dates, since a thing ships when it works on a real phone and a real computer.}
  \label{fig:roadmap}
\end{figure}

\nanomuse is a non-profit project that would like a community; this roadmap is an offer of work in three distances, and \cref{fig:roadmap} lays it out as five lanes, each Now card with a first task.

\bpar{Now: catching up with \muse.} The near work is unglamorous, and the first column of \cref{fig:roadmap} lists it: memory that surfaces at the right moment, proactivity between the schedules, a conversation in which a person who has used \muse does not feel the drop, hands that fail less, self-hosting in one command.

\bpar{Next.} An evaluation suite built from failed runs, so that each release has a number rather than an impression, run beside the public suites: AndroidWorld for the phone~\cite{rawles2024androidworld}, OSWorld for the desktop~\cite{xie2024osworld}, MemGUI-Bench for what the hands remember across apps~\cite{liu2026memgui} and OS-Harm for what they should refuse~\cite{kuntz2025osharm}. Then a small open model of our own for the hands, trained on traces people chose to contribute, as MobileForge adapts an open model to real apps from runs it collected itself~\cite{liu2026mobileforge}, so that the most sensitive loop does not depend on a vendor.

\bpar{Later: looking up.} The physical world as hands: the hub's frames carry look, act and ask already, so a camera, an appliance or a robot arm is one more device, under the same approvals and the same log; \muse's gadget SDK is a vendor's first shape of this (\cref{sec:muse}). Beyond that, an agent that lasts years, with memory and identity that survive models, devices and vendors.

\bpar{How to join.} The repository and the project's chat server are the way in. A skill, a translation, a device, a failing trace of the hands, a wrong paragraph in this report: each is a contribution, and the small ones are the ones we need most.

%% file: sec/7_limitations.tex
\section{Limitations}
\label{sec:limits}

\bpar{A policy boundary is not a privilege boundary.} On a single device the \sentinel runs in the same trust domain as the agent. A persuaded model can be stopped by the rules, and the rules cannot be edited by the model, but there is no host outside the agent's reach holding the real credentials, as there is in \muse; a compromise of the device is a compromise of the agent. The decision order, the taint rule and the Linux sandbox mitigate this; they do not solve it.

\bpar{The hands are judged by their words, and not yet measured.} A button that is only an icon, in an app that labels nothing, is judged by neither the label nor the accessibility text, and on a dense professional screen even finding it is unsolved~\cite{li2025screenspotpro}; the person watching the ring is the last line. We have no success rate and no count of human take-overs to report: the number does not exist yet, and we would rather say so than estimate it. On Linux the hands work in X11 sessions only.

\bpar{Memory does not yet know who wrote it.} A line a weak model guessed is read as fact by a stronger one; the store records when a line changed and lets the person undo it, but not the model, the date or a confidence, and there is no rule for re-checking an old line. Until there is, the memory page should be read as the notes of a colleague one does not yet fully trust.

\bpar{The relay is shared, and it keeps text.} A person who syncs their conversations is storing their text on a server they do not run, and the community relay is one process on one machine operated by the project; we have made the shared part small and runnable at home, not unnecessary. The default-on switch for improving the community's models favours an open model over the quietest default, and it is the choice in this project we are least sure of.

\bpar{Where the record is thin.} Our account of \muse rests on four public documents, the shipped clients, and a copy of the production system prompt whose provenance we cannot verify beyond its consistency with the other two; prompts change with every deployment, and what we read in October 2026 may not be what runs when this report is read. A statement marked ``prompt'' or ``inferred'' should be weighed accordingly; where Meta publishes more we will correct it.

\section*{Acknowledgements}
\phantomsection\addcontentsline{toc}{section}{Acknowledgements}% the anchor sits on the heading's page, not the previous one

\nanomuse stands on other people's work: OpenMinis for the whole on-device construction~\cite{openminis2026}, the DeepSeek harness for the desktop~\cite{deepseek-harness2026}, UI-TARS-desktop~\cite{uitarsdesktop2025} and MemGUI-Bench~\cite{liu2026memgui} for the operators, \muse's designers and its safety engineer for the record of what they built. The early users, the senders of failing traces and the contributors made it what it is.